# Color Image Compression Algorithm Based on the DCT Blocks

*Walaa M. Abd-Elhafiez, **Wajeb Gharibi

*Mathematical Department, Faculty of Science, Sohag University, 82524, Sohag, Egypt.
*,**College of Computer Science & Information Systems, Jazan University, Jazan, Kingdom of Saudi Arabia.

*Abstract*— This paper presents the performance of different block-based discrete cosine transform (DCT) algorithms for compressing color image. In this RGB component of color image are converted to YCbCr before DCT transform is applied. Y is luminance component; Cb and Cr are chrominance components of the image. The modification of the image data is done based on the classification of image blocks to edge blocks and non-edge blocks, then the edge block of the image is compressed with low compression and the non-edge blocks is compressed with high compression. The analysis results have indicated that the performance of the suggested method is much better, where the constructed images are less distorted and compressed with higher factor.

## I. Introduction

The objective of an image compression technique is to represent an image with smaller number of bits without introducing appreciable degradation of visual quality of decompressed image. These two goals are mutually conflict in nature. In a digital true color image, each color component that is R, G, B components, each contains 8 bits data [1]. Also color image usually contains a lot of data redundancy and requires a large amount of storage space. In order to lower the transmission and storage cost, image compression is desired [2]. Most color images are recorded in RGB model, which is the most well known color model. However, RGB model is not suited for image processing purpose. For compression, a luminance-chrominance representation is considered superior to the RGB representation. Therefore, RGB images are transformed to one of the luminance-chrominance models, performing the compression process, and then transform back to RGB model because displays are most often provided output image with direct RGB model. The luminance component represents the intensity of the image and look likes a gray scale version. The chrominance components represent the color information in the image [3,4].

Douak et al. [5] have proposed a new algorithm for color images compression. After a preprocessing step, the DCT transform is applied and followed by an iterative phase including the threshold, the quantization, dequantization and the inverse DCT. For the aim, to obtain the best possible compression ratio, the next step is the application of a proposed adaptive scanning providing, for each (n, n) DCT block a corresponding (n×n) vector containing the maximum possible run of zeros at its end. The last step is the application of a modified systematic lossless encoder. The efficiency of their proposed scheme is demonstrated by results.

Mohamed et al. [6] proposed a hybrid image compression method, which the background of the image is compressed using lossy compression and the rest of the image is compressed using lossless compression. In hybrid compression of color images with larger trivial background by histogram segmentation, input color image is subjected to binary segmentation using histogram to detect the background. The color image is compressed by standard lossy compression method. The difference between the lossy image and the original image is computed and is called as residue. The residue at the background area is dropped and rest of the area is compressed by standard lossless compression method. This method gives lower bit rate than the lossless compression methods and is well suited to any color image with larger trivial background.

In this paper, the proposed method for color image compression makes a balance on compression ratio and image quality by compressing the vital parts of the image with high quality. In this approach the main subject in the image is very important than the background image. Considering the importance of image components and the effect of smoothness in image compression, this method classifies the image as edge blocks (main subject) and non-edge blocks (background), then the background of the image is subjected to low quality lossy compression and the main subject is compressed with high quality lossy compression. We tested our algorithm with different kind of image and the experimental results show the effectiveness of our approach. The rest of our paper is organized as follow, The JPEG Compression method and detect edges used in the proposed work are described in section II. Section III describes the proposed algorithm. Section IV presents the experimental results obtained in this paper. Section V draws the conclusion of this work.

## II. Background

### A. JPEG Compression

*Components of Image Compression (JPEG) System.* Image compression system consists of three closely connected components namely (Source encoder (DCT based), Quantizer and Entropy encoder). Figure 1(a) shows the architecture of the JPEG encoder.

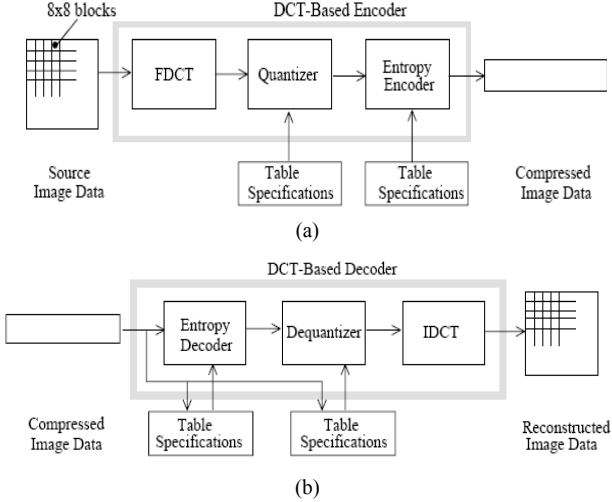

Figure 1. Typical image coding system (JPEG encoder/decoder), a) Block diagram of JPEG encoder processing steps, b) Block diagram of JPEG decoder.

*Principles behind JPEG Compression*, a common characteristic of most images is that the neighboring pixels are correlated and therefore contain redundant information. The foremost task then is to find less correlated representation of the image. Two fundamental components of compression are redundancy and irrelevancy reduction. Redundancy reduction aims at removing duplication from the signal source. Irrelevancy reduction omits parts of the signal that will not be noticed by the signal receiver, namely the Human Visual System (HVS). The JPEG compression standard (DCT based) employs the use of the discrete cosine transform, which is applied to each 8x8 block of the partitioned image. Compression is then achieved by performing quantization of each of those 8x8 coefficient blocks.

Image transform coding for JPEG compression algorithm. In the image compression algorithm, the input image is divided into 8-by-8 or 16-by-16 non-overlapping blocks, and the two-dimensional DCT is computed for each block. The DCT coefficients are then quantized, coded, and transmitted. The JPEG receiver (or JPEG file reader) decodes the quantized DCT coefficients, computes the inverse two-dimensional DCT of each block, and then puts the blocks back together into a single image. For typical images, many of the DCT coefficients have values close to zero; these coefficients can be discarded without seriously affecting the quality of the reconstructed image. A two dimensional DCT of an N by N matrix pixel is defined as follows

$$DCT(i, j) = \frac{1}{\sqrt{2N}} C(i)C(j) \sum_{x=0}^{N-1} \sum_{y=0}^{N-1} pixel(x, y) \cos\left[\frac{(2x+1)i\pi}{2N}\right] \cos\left[\frac{(2y+1)j\pi}{2N}\right]$$

where $C(x) = \begin{cases} \frac{1}{\sqrt{2}} & \text{if } x = 0 \\ 1 & \text{otherwise} \end{cases}$

For decoding purpose there is an inverse DCT (IDCT):

$$pixel(x, y) = \frac{1}{\sqrt{2N}} \sum_{i=0}^{N-1} \sum_{j=0}^{N-1} C(i)C(j)DCT(i, j) \cos\left[\frac{(2x+1)i\pi}{2N}\right] \cos\left[\frac{(2y+1)j\pi}{2N}\right]$$

The DCT based encoder can be thought of as essentially compression of a stream of 8x8 blocks of image samples. Each 8 X 8 block makes its way through each processing step, and yields output in compressed form into the data stream. Because adjacent image pixels are highly correlated, the forward DCT (FDCT) processing step lays the foundation for achieving data compression by concentrating most of the signal in the lower spatial frequencies. For a typical 8x8 sample block from a typical source image, most of the spatial frequencies have zero or near-zero amplitude and need not be encoded. In principle, the DCT introduces no loss to the source image samples; it merely transforms them to a domain in which they can be more efficiently encoded. After output from the FDCT, each of the 64 DCT coefficients is uniformly quantized in conjunction with a carefully designed 64–element quantization table (QT). At the decoder, the quantized values are multiplied by the corresponding QT elements to recover the original unquantized values. After quantization, all of the quantized coefficients are ordered into the "zig-zag" sequence as shown in figure 2.

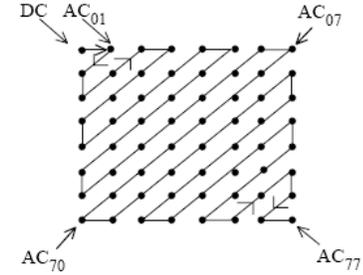

Fig 2: Zig Zag Sequence

This ordering helps to facilitate entropy encoding by placing low-frequency non-zero coefficients before high-frequency coefficients. The DC coefficient, which contains a significant fraction of the total image energy, is differentially encoded. Figure 1 (b) show the JPEG decoder architecture, which is the reverse procedure described for compression.

### B. Edge Detection

There are several techniques have been used for edge detection [8]. In this paper, Canny Method is used, the canny edge detection algorithm is known to many as the optimal edge detector [9]. Firstly it smoothes the image to eliminate the noise. It then finds the image gradient to highlight regions with high spatial derivatives. The algorithm then tracks along these regions and suppresses any pixel that is not at the maximum (nonmaximum suppression). The gradient array is now further reduced by hysteresis. Hysteresis is used to track along the remaining pixels that have not been suppressed. Hysteresis uses two thresholds and if the magnitude is below the first threshold, it is set to zero (made a nonedge). If the magnitude is above the high threshold, it is made an edge. And if the magnitude is between the two thresholds, then it is set to zero unless there is a path from this pixel to a pixel with a gradient above the second threshold.

### III. The Proposed Image Coding Scheme

The proposed technique built around several steps. Each step will be explained in more details in the following:

*a) Transform RGB to YCbCr[11]*

Before the application of the RGB to YCbCr transformation, the mean values of the three plane images R, G and B are removed. The use of the already reported transformation is due to the fact that almost signal energy of the new transformed YCbCr image is contained in the Y plane. Consequently, we can achieve an efficient compression that allows reaching high compression ratios in the Cb and Cr without loosing the quality of the whole compressed image when returned to the original RGB space. It means that acting on YCbCr space proffers better performances than the original RGB space. The RGB to YCbCr is performed respecting to

$$\begin{bmatrix} Y \\ Cb \\ Cr \end{bmatrix} = \begin{bmatrix} 0.299 & 0.587 & 0.144 \\ -0.16875 & -0.33126 & 0.5 \\ 0.5 & -0.41869 & -0.08131 \end{bmatrix} \begin{bmatrix} R \\ G \\ B \end{bmatrix}$$

However, the inverse transformation is simply expressed by

$$\begin{bmatrix} R \\ G \\ B \end{bmatrix} = \begin{bmatrix} 1 & 0 & 1.402 \\ 1 & -0.34413 & -0.71414 \\ 1 & 1.772 & 0 \end{bmatrix} \begin{bmatrix} Y \\ Cb \\ Cr \end{bmatrix}$$

*b) Block-based DCT transform*

For any color image, after the preprocessing step (means removing and RGB to YCbCr), each one of the new three planes (Y,Cb,Cr) are partitioned to blocks and classification the blocks into edge (foreground and more important block) and non-edge (background and less important block). The classification process is accomplished by using canny method. The different block sizes: 8×8, 16×16 or 32×32 were tested. Each block is DCT transformed. It is clear, that DCT transform (such as the wavelets) concentrate the great part of block energy in few representative coefficients.

The DCT coefficients in each block are then uniformly quantized with quantization step sizes depending on the DCT coefficient. The step sizes are represented in a quantization matrix called the Q-matrix. Different Q-matrices are typically used for the luminance and chrominance components. This quantization stage determines both the amount of compression and the quality of the decompressed image. Large quantization step sizes give good compression but poor visual performance while small quantization step sizes give good visual performance but small compression.

In the first scheme (M-1), all AC coefficients of the edge blocks on each components (Y, Cb and Cr) are used. After quantization and zigzag scan the non-zero of the quantized coefficients is counted and all AC coefficients will be used as the input of the huffman coding. The non-edge block will be coded using only the DC coefficient. The results of the M-1 scheme are given in Table 1.

In the second scheme (M-2), a 70% of the non-zero AC coefficients of the edge blocks on each components provides good results. After quantization and zigzag scan the non-zero of the quantized coefficients is counted and only the first 70% of the non-zero AC coefficients on each component will be used as the input of the huffman coding. The non-edge block will be coded using only the DC coefficient. The results of the M-2 method are given in Table 1.

In the third technique (M-3), a 50% of the non-zero AC coefficients of the edge blocks on Y component, 50% of the non-zero AC coefficients of the edge blocks on Cb component, and 50% of the non-zero AC coefficients of the edge blocks on Cr component provides an accepted results. After quantization and zigzag scan the non-zero of the quantized coefficients is counted and only the first 50% of the non-zero AC coefficients on each component will be used as the input of the huffman coding. The non-edge block will be coded using only the DC coefficient. The results of the M-3 scheme are given in Table 1. M-3 scheme provides improvement in the CR from 9.35 to 14.61 relative to the M-1 scheme (at 8×8 block size) with a little decreasing of the image quality and PSNR.

### IV. Experimental Results

The proposed image coding scheme is implemented according to the description in section III and tested with different block size (8×8, 16×16 and 32×32) and with a set of test color images shown in figure 3 (Nile of size 512×512, and Barbara, Gold, Couple and Girl of size 256×256).

Here compression ratio is measured in terms of bpp and the image quality in terms of PSNR and visual fidelity index [10]. The bpp and PSNR may be defined, respectively, as

$$\text{bpp} = \frac{\text{size of compressed color image in bits}}{\text{number of pixels}}$$

and

$$\text{PSNR} = 10 \times \log_{10} \frac{255^2 \times 3}{\text{MSE(Y)} + \text{MSE(Cb)} + \text{MSE(Cr)}}$$

where MSE is the mean square error for each space.

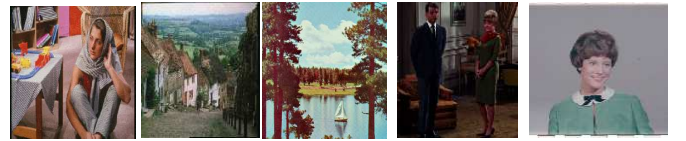

Fig. 3. Original test images: (a) Barbara, (b) Gold, (c) Nile, (d) Couple and (e) Girl.

Fig. 4, Fig. 5 and Fig. 6 give visual and quantitative results of the method considering the RGB to YCbCr transformation for block size 8×8, 16×16 and 32×32, respectively. Table 2, table 3 and table 4 show the comparison of the results with the proposed technique to the CBDCT-CABS [5], RGB space method [7] and JPEG image compression schemes.

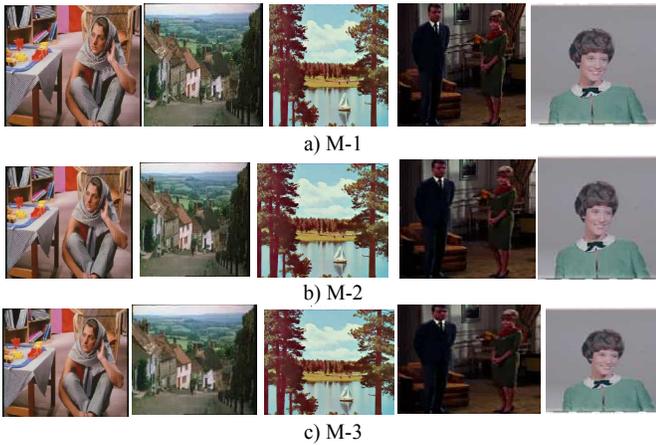

a) M-1

b) M-2

c) M-3

Fig. 4. Compressed images visual performance of the M-1, M-2 and M-3 techniques (DCT block size is **8×8**).

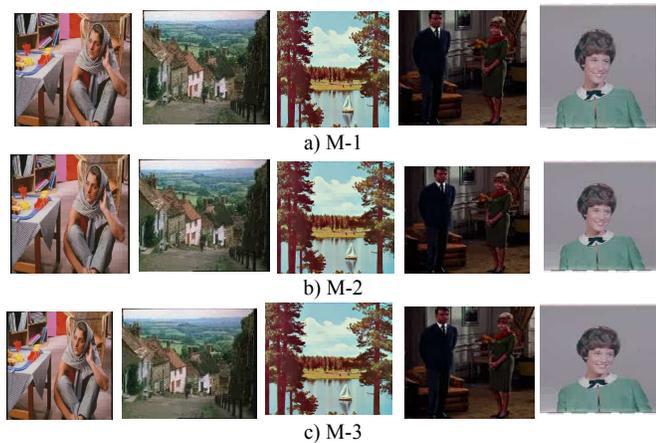

a) M-1

b) M-2

c) M-3

Fig. 4. Compressed images visual performance of the M-1, M-2 and M-3 techniques (DCT block size is **16×16**).

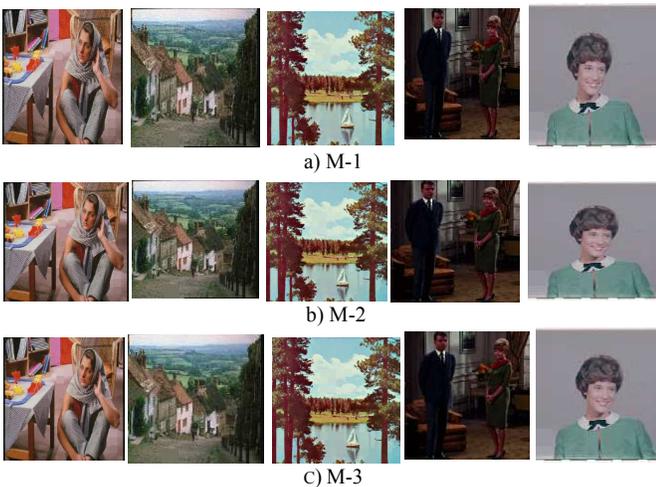

a) M-1

b) M-2

c) M-3

Fig. 5. Compressed images visual performance of the M-1, M-2 and M-3 techniques (DCT block size is **32×32**).

The average of the proposed method (M-3) with little bit rate give better PSNR than recent publisher (CBDCT-CABS [5], RGB space method [7]) methods and high compression as shown in table 2 and table 3. The results of the proposed method (M-3) show an improvement over JPEG, for different types of images. The proposed method's (M-3) PSNR is higher than that of JPEG by approximately 2.22 dB.

Table 2: Comparison between the CBDCT-CABS [5] algorithm and the proposed method (M-3).

|  | CBDCT-CABS [5] | | | Proposed method (M-3) | | |
|---|---|---|---|---|---|---|
| image | PSNR | bpp | CR | PSNR | bpp | CR |
| **Block Size (16×16)** | | | | | | |
| lena | 31.94 | 0.8625 | 27.826 | 33.1879 | 0.3506 | 68.4501 |
| Airplane | 30.251 | 0.6247 | 38.417 | 34.0504 | 0.3872 | 61.9830 |
| Fruit | 30.189 | 0.899 | 26.698 | 37.5089 | 0.2958 | 81.1466 |
| Average | 30.793 | 0.7954 | 30.980 | **34.9157** | **0.3445** | **70.526** |
| **Block Size (32×32)** | | | | | | |
| lena | 31.837 | 0.7865 | 30.516 | 32.8415 | 0.2458 | 97.6540 |
| Airplane | 30.346 | 0.5786 | 41.481 | 33.7926 | 0.3157 | 76.0334 |
| Fruit | 30.147 | 0.9192 | 26.111 | 33.4935 | 0.5184 | 46.2968 |
| Average | 30.77667 | 0.761433 | 32.70267 | **33.3758** | **0.3599** | **73.328** |

Table 3: Comparison between the proposed method (M-1) and the RGB space method [7].

|  | RGB space method [7] | | | Proposed method (M-1) | | |
|---|---|---|---|---|---|---|
| image | PSNR | bpp | CR | PSNR | bpp | CR |
| **Block Size (16×16)** | | | | | | |
| lena | 35.477 | 1.114 | 21.53 | 34.5728 | 0.7125 | 33.6831 |
| Airplane | 35.472 | 0.871 | 27.53 | 35.8720 | 0.8095 | 29.6494 |
| Fruit | 34.382 | 0.865 | 27.73 | 33.2782 | 0.6278 | 38.2311 |
| Average | 35.110 | 0.950 | 25.60 | **34.57433** | **0.7166** | **33.854** |
| **Block Size (32×32)** | | | | | | |
| lena | 35.698 | 1.178 | 20.35 | 34.1700 | 0.6740 | 35.6081 |
| Airplane | 35.488 | 0.846 | 28.36 | 35.3103 | 0.8142 | 29.4772 |
| Fruit | 34.851 | 1.079 | 22.24 | 32.4248 | 0.4786 | 50.1431 |
| Average | 35.346 | 1.034 | 23.65 | **33.96837** | **0.6556** | **38.409** |

Table 4: Comparative results of the proposed method (M-3) and JPEG method with block size (8×8).

|  | JPEG | | Proposed method (M-3) | |
|---|---|---|---|---|
| image | PSNR | bpp | PSNR | bpp |
| lena | 32.76 | 1.03 | 33.9259 | 0.6587 |
| Fruit | 30.47 | 1.47 | 33.3902 | 0.7342 |
| Airplane | 31.46 | 0.90 | 34.3943 | 0.6351 |
| House | 31.34 | 1.24 | 33.5263 | 0.6578 |
| Zelda | 32.06 | 1.00 | 33.9442 | 0.6321 |

## V. Conclusion

In this paper a block-based coding scheme was proposed along with its applications to compress color images. The obtained results shows the improvement of the proposed method over the recent published paper both in quantitative PSNR terms and very particularly, in visual quality of the reconstructed images. Furthermore, it increased the compression rate.

## References

[1] Rafael C. Gonzalez and Richard E. Woods. Digital Image Processing. Pearson Education, Englewood Cliffs,2002 .

[2] C. Yang, J. Lin, and W. Tsai, "Color Image Compression by Moment-preserving and Block Truncation Coding Techniques", IEEE Trans.Commun., vol.45, no.12,pp.1513-1516, 1997.

[3] S. J. Sangwine and R.E.Horne, "The Colour Image Processing Handbook", Chapman & Hall, 1st Ed., 1998.

[4] M. Sonka, V. Halva, and T.Boyle, "Image Processing Analysis and Machine Vision", Brooks/Cole Publishing Company, 2nd Ed., 1999.

[5] F. Douak, Redha Benzid, Nabil Benoudjit "Color image compression algorithm based on the DCT transform combined to an adaptive block scanning," Int. J. Electron. Commun. (AEU), vol. 65, pp. 16–26, 2011.

[6] M. Mohamed Sathik, K.Senthamarai Kannan and Y.Jacob Vetha Raj, "Hybrid Compression of Color Images with LargerTrivial Background by Histogram Segmentation", (IJCSIS) International Journal of Computer Science and Information Security, Vol. 8, No. 9, December 2010.

[7] Walaa M. Abd-Elhafiez, "New Approach for Color Image Compression", International Journal of Computer Science and Telecommunications (IJCST), Volume 3, Issue 4, pp. 14-19, April 2012.

[8] Davis. L, "Survey of edge detection techniques, computer vision," *Graph. Image Process*, vol. 4, pp. 248–270, 1975.

[9] Canny. J, "A computational approach to edge detection," *IEEE Trans. Pattern Anal. Machine. Intell.*, vol. PAMI-8, pp. 679–698, Nov. 1986.

[10] A. Toet, M.P. Lucassen, "A new universal colour image fidelity metric", Displays 24, pp. 197–207, 2003.

[11] Arash Abadpour, Shohreh Kasaei, "Color PCA Eigenimages and their Application to Compression and Watermarking", submitted to Image & Vision Computing 21 August 2007.

Table 1. Performances in the YCbCr space for the different DCT block sizes.

| IMAGE | M-1 | | | M-2 | | | M-3 | | |
|---|---|---|---|---|---|---|---|---|---|
| | bpp | PSNR | CR | bpp | PSNR | CR | bpp | PSNR | CR |
| **BLOCK SIZE (8×8)** | | | | | | | | | |
| Barbara512 | 1.2555 | 33.8035 | 19.1163 | 0.8963 | 32.1209 | 26.7779 | 0.7444 | 31.6005 | 32.2402 |
| gold256 | 1.3811 | 32.9164 | 17.3780 | 0.9630 | 31.7899 | 24.9234 | 0.7931 | 31.3838 | 30.2602 |
| Nile | 1.4196 | 33.1968 | 16.9065 | 1.0483 | 32.3825 | 22.8949 | 0.8627 | 32.0397 | 27.8202 |
| Couple | 0.9839 | 36.6389 | 24.3931 | 0.6933 | 34.8427 | 34.6187 | 0.5732 | 34.1120 | 41.8716 |
| Girl | 0.6220 | 39.7067 | 38.5837 | 0.4680 | 37.4917 | 51.2868 | 0.3969 | 36.7575 | 60.4622 |
| **Average** | **1.1324** | **35.2524** | **23.27552** | **0.8137** | **33.7255** | **32.10034** | **0.6740** | **33.1787** | **38.53088** |
| **BLOCK SIZE (16×16)** | | | | | | | | | |
| barbara256 | 2.1921 | 36.9712 | 10.9484 | 1.5432 | 34.6936 | 15.5517 | 1.1412 | 33.2037 | 21.0298 |
| gold256 | 2.7010 | 37.6028 | 8.8854 | 1.8789 | 34.3976 | 12.7734 | 1.3945 | 33.1004 | 17.2108 |
| Nile | 3.9608 | 35.3381 | 6.0594 | 2.5395 | 34.1420 | 9.4508 | 1.8039 | 33.3714 | 13.3047 |
| Couple | 2.0086 | 38.3494 | 11.9485 | 1.1667 | 37.0057 | 20.5705 | 0.8421 | 35.7510 | 28.4986 |
| Girl | 0.9462 | 41.5581 | 25.3659 | 0.6040 | 39.5123 | 39.7318 | 0.4550 | 38.2718 | 52.7435 |
| **Average** | **2.3617** | **37.9639** | **12.64152** | **1.5464** | **35.9502** | **19.61564** | **1.1273** | **34.7396** | **26.55748** |
| **BLOCK SIZE (32×32)** | | | | | | | | | |
| barbara256 | 2.3739 | 37.9212 | 10.1098 | 1.6272 | 35.2770 | 14.7494 | 1.1787 | 33.5831 | 20.3618 |
| gold256 | 2.7692 | 38.0313 | 8.6668 | 1.9036 | 34.7695 | 12.6078 | 1.3871 | 33.3982 | 17.3017 |
| Nile | 4.4459 | 35.5570 | 5.3982 | 2.6110 | 34.2335 | 9.1920 | 1.8000 | 33.4316 | 13.3335 |
| Couple | 2.4588 | 38.4064 | 9.7610 | 1.1734 | 37.1074 | 20.4536 | 0.8089 | 35.8671 | 29.6683 |
| Girl | 1.1430 | 40.7595 | 20.9970 | 0.6310 | 39.1959 | 38.0360 | 0.4504 | 38.0046 | 53.2831 |
| **Average** | **2.6381** | **38.1350** | **10.98656** | **1.5892** | **36.1166** | **19.00776** | **1.1250** | **34.8569** | **26.78968** |